\documentclass[letterpaper, 10 pt, conference]{ieeeconf}  

\IEEEoverridecommandlockouts                              

\usepackage{graphics} 
\usepackage{amsmath} 
\usepackage{amssymb}  
\usepackage[backgroundcolor=cyan]{todonotes}

\usepackage{algorithm, algorithmic}
\usepackage{microtype}

\newcommand{\myvec}[1]{\boldsymbol{#1}}

\newcommand{\vla}{\mbox{\boldmath $\lambda$}}

\newcommand{\vta}{\mbox{\boldmath $\tau$}}

\newcommand{\vom}{\mbox{\boldmath $\omega$}}

\newcommand{\vc}{\myvec{c}}

\newcommand{\vg}{\myvec{g}}

\newcommand{\vp}{\myvec{p}}
\newcommand{\vq}{\myvec{q}}

\newcommand{\vu}{\myvec{u}}

\newcommand{\vx}{\myvec{x}}

\newcommand{\vA}{\myvec{A}}
\newcommand{\vB}{\myvec{B}}
\newcommand{\vC}{\myvec{C}}

\newcommand{\vG}{\myvec{G}}

\newcommand{\vJ}{\myvec{J}}
\newcommand{\vK}{\myvec{K}}

\newcommand{\vM}{\myvec{M}}

\newcommand{\vR}{\myvec{R}}
\newcommand{\vS}{\myvec{S}}
\newcommand{\vT}{\myvec{T}}
\newcommand{\vU}{\myvec{U}}

\newcommand{\vX}{\myvec{X}}

\DeclareMathOperator{\sig}{sig}

\usepackage{cite} 

\title{\LARGE \bf
Whole-Body Nonlinear Model Predictive Control Through Contacts for Quadrupeds
}

\author{
Michael Neunert$^{1}$, 
Markus St{\"a}uble$^{1}$,
Markus Giftthaler$^{1}$,
Carmine D. Bellicoso$^{2}$,
Jan Carius$^{2}$ \\
Christian Gehring$^{2}$,
Marco Hutter$^{2}$
and Jonas Buchli$^{1}$
\thanks{
$^1$Agile \& Dexterous Robotics Lab, ETH Z\"urich, Switzerland. {\small \{neunertm, mgiftthaler, markusta, buchlij\}@ethz.ch} \newline
$^2$Robotic Systems Lab, ETH Z\"urich, Switzerland. {\small \{bcarmine, jcarius, gehrinch, mahutter\}@ethz.ch }
}
}

\begin{document}
\maketitle
\thispagestyle{empty}
\pagestyle{empty}

\begin{abstract}
In this work we present a whole-body Nonlinear Model Predictive Control approach for Rigid Body Systems subject to contacts. We use a full dynamic system model which also includes explicit contact dynamics. Therefore, contact locations, sequences and timings are not prespecified but optimized by the solver. Yet, thorough numerical and software engineering allows for running the nonlinear Optimal Control solver at rates up to 190 Hz on a quadruped for a time horizon of half a second. This outperforms the state of the art by at least one order of magnitude. Hardware experiments in form of periodic and non-periodic tasks are applied to two quadrupeds with different actuation systems. The obtained results underline the performance, transferability and robustness of the approach.

\end{abstract}

\section{INTRODUCTION}
In this paper, we present a whole-body Nonlinear Model Predictive Control (NMPC) approach for Rigid Body Dynamics (RBD) systems subject to contacts. By using an explicit, Auto-Differentiable contact model as part of the dynamic system, the approach is able to reason about contacts and optimize through them efficiently. Contact timings, sequences or locations are not pre-specified, but an outcome of the optimization. Thanks to a highly-efficient, unconstrained nonlinear optimal control solver, we are able to successfully apply the method to two different quadruped platforms. We verify the performance and versatility of our approach by testing a multitude of tasks including periodic gait patterns as well as highly dynamic motions, such as squat jumps.

\subsection{Related Work}
Motion planning and control for legged, and especially quadruped locomotion is often tackled with multi-stage planning and control frameworks \cite{Winkler2015,littledog,Gehring2016, mastalli2016}. Such frameworks often consist of one or multiple planner stages that use a simplified model and a reactive tracking controller. This results in the dilemma that the planner does not always produce feasible, i.e. dynamically consistent, plans or needs to plan conservatively. The tracking controller on the other hand does not have enough control authority to e.g. modify foothold positions or contact timings, but blindly tries to track the planners reference. In this field, centroidal dynamics approaches \cite{Dai2015,Kuindersma2016,koyanagi,farshidian_locomotion, herzog_centroidal} become increasingly popular as they capture the core dynamics of the problem. However, many of these approaches plan or optimize contact forces that are not guaranteed to be realizable.

In recent years, there has been an increasing number of whole-body optimization and optimal control based approaches \cite{mombaur2009, Posa2013, Tassa2012, pardo_projection, mastalli2016}. While these approaches are very complete, their runtimes are still a few orders of magnitudes away from running in receding horizon or MPC fashion. There are also some whole-body NMPC approaches verified in simulation \cite{Erez2013}. However, without hardware validation, it remains an open question if and how well these approaches can be applied to real robots. While whole-body, contact invariant NMPC has been demonstrated on hardware before~\cite{koenemann2015whole}, the presented motions were rather slow or even quasi static, underlined by the fact that the authors do not apply the torque output to the robot but instead only use the position and velocity trajectories as references for the joint controllers. Additionally, contact switching was not dynamic and artificially enforced. Also stability during execution was explicitly encoded in the task.
\begin{figure}
\centering
\includegraphics[width=0.445\columnwidth]{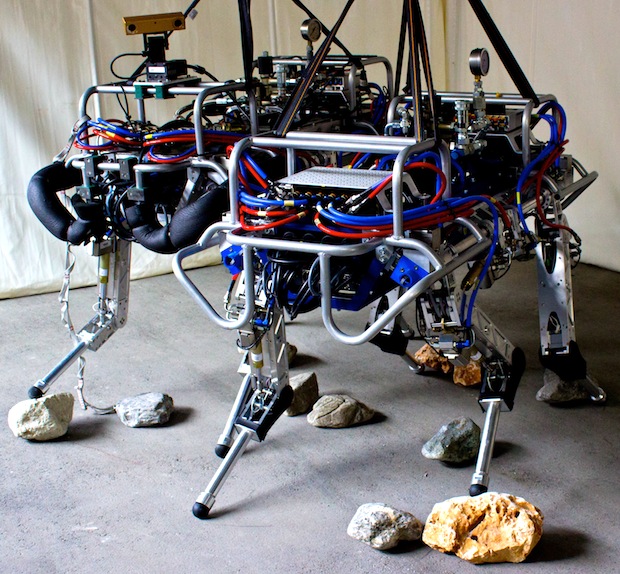}
\includegraphics[width=0.45\columnwidth]{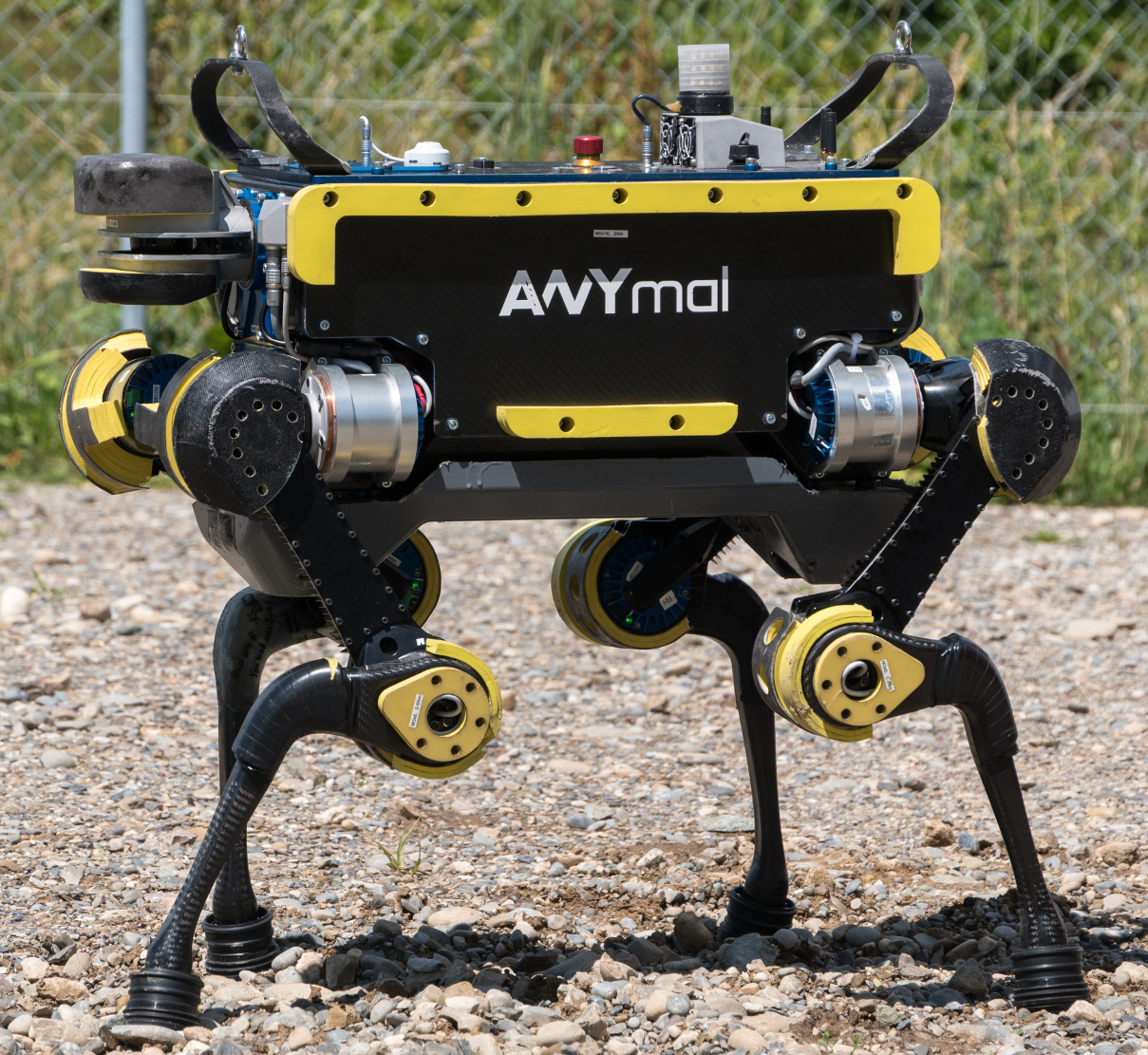}
\footnotesize
\caption{The quadrupeds HyQ-blue (left front) and ANYmal (right), which served as test platforms for our MPC experiments.}
\label{fig:hyq_and_anymal}
\end{figure}

\subsection{Contributions}
In this work, we demonstrate whole-body, contact invariant nonlinear MPC for highly dynamic motions that require explicit reasoning about the full dynamics of the system and the contacts. Furthermore, to the best of our knowledge, we demonstrate that such a framework can be applied to both single motions as well as periodic gaits on hardware for the first time. We show that the approach transfers between platforms, and apply the same framework to the two quadrupeds HyQ and ANYmal (Figure~\ref{fig:hyq_and_anymal}). We also demonstrate the robustness and replanning capabilities of the approach by adding significant disturbances during execution. We summarize our solver framework, which uses Auto-Differentiation and code generation to achieve high computational performance exceeding the current state of the art in robotics NMPC applications by at least one order of magnitude. In contrast to many previous approaches, our solver is also available as open-source software~\cite{adrlCT}. Furthermore, we also publish the cost function weights used during experiments to ensure reproducibility.

In previous work \cite{neunert:2017:ral}, we demonstrated the versatility of optimizing whole-body motions through contacts. However, trajectories were only optimized once before execution. Also, especially interesting tasks such as periodic gaits could not be transferred to hardware due to model mismatches and lack of robustness of the plans. In this work, we demonstrate that an NMPC approach which continuously re-optimizes the state and control trajectories at high frequency, results in robust performance and copes with model mismatches. Running NMPC on real hardware poses severe restrictions in terms of computation time and software integration. In this work, we describe how we overcome these issues and improve our solver to achieve a speedup of several orders of magnitude.

\subsection{Structure of this paper}
We organize the paper as follows. In Section~\ref{sec:nmpc_for_rbd}, we define how we formulate the NMPC problem for Rigid Body Dynamics Systems. In Section~\ref{sub:solver} we describe our approach of solving the problem. Afterwards, in Sections~\ref{sec:software} and \ref{sec:hardware_integration}, we describe the implementation from a software point of view as well as the integration on hardware. In Section~\ref{sec:results}, we then present our experiments and discuss the results and findings. Finally, in Section~\ref{sec:summary} we summarize the paper and provide an outlook to future work.

\section{NMPC FOR RIGID BODY SYSTEMS} \label{sec:nmpc_for_rbd}
The NMPC approach used in this work recurrently solves finite-horizon Optimal Control Problems with cost functions of the form
\begin{align}
J(\vx(t),\vu(t)) = h\left( \vx(t_f) \right) + \int_{t=0}^{t_f} L \left(  
\vx(t),\vu(t), t
\right) dt
\label{eq:cost_function}
\end{align}
and nonlinear system dynamics
\begin{align}
\dot{\vx}(t) = f(\vx(t),\vu(t), t), \quad \vx(0) = \vx_0
\label{eq:dynamic_system}
\end{align}
with state and control trajectories $\vx(t)$ and $\vu(t)$.

\subsection{System Modelling}
Analogously to~\cite{neunert:2017:ral}, we consider general Rigid Body Dynamics of the form
\begin{equation}
    \vM(\vq)\ddot{\vq} + \vC(\vq,\dot{\vq}) + \vG(\vq) = \vJ^\top_c \vla (\vq, \dot{\vq}) + \vS^\top \vta
    \label{eq:rbd}
\end{equation}
where $\vM_{18\times 18}$ is the inertia matrix,~$\vC_{18\times 1}$ captures Coriolis and centripetal forces and $\vG_{18\times 1}$ represents the gravity terms. We further assume the system is actuated via joint forces/torques~$\vta_{12\times 1}$ where the selection matrix~$\vS_{18 \times 12}$ maps these forces to the actuated degrees of freedom. Additionally, external forces $\vla_{12\times 1}$ act on the system via the contact Jacobian~$(\vJ_c)_{18\times 12}$. The joint positions/angles and their velocities are denoted by~$\vq$ and~$\dot{\vq}$ respectively. In the case of a floating base robot, these quantities also contain the base pose (i.e. its orientation and position) as well as the base twist (i.e. its linear and angular velocity). These quantities can also be thought of as an unactuated 6~DoF joint between an inertial and the base frame. We use an Euler-Angle parametrization of the 3D orientation.

We define the state of our system as follows
\begin{equation}
    \vx = [{}_{\mathcal{W}}\vq^\top~{}_{\mathcal{L}}\dot{\vq}^\top]^\top = [{}_{\mathcal{W}}\vq_B^\top~{}_{\mathcal{W}}\vx_B^\top~\vq_J^\top~{}_{\mathcal{L}}\vom_B^\top~{}_{\mathcal{L}}\dot{\vx}_B^\top~\dot{\vq}_J^\top]^\top
    \label{eq:coordinates}
\end{equation}
where the pose is expressed in an inertial ``world'' frame $\mathcal{W}$ and the twist is expressed in a local body frame $\mathcal{B}$. Due to the different reference frames, the twist is not a pure time derivative of the pose, but requires an additional coordinate transform $\vT_{\mathcal{W}\mathcal{L}}$ which is composed of a pure rotation matrix for linear velocities as well as a slightly more complex mapping matrix for angular velocities. This leads to the overall system dynamics
\begin{align}
\dot{\vx}(t) &= \left[ {}_{\mathcal{W}}\dot{\vq}^\top \quad {}_{\mathcal{L}}\ddot{\vq}^\top \right]^\top
             \\ &= 
             \begin{bmatrix}
                \vT_{\mathcal{W}\mathcal{L}}~{}_{\mathcal{L}}\dot{\vq} \\
                \vM^{-1}(\vq)(\vS^\top\vta + \vJ_c \vla(\vq,\dot{\vq}) - \vC(\vq, \dot{\vq}) - \vG(\vq))
             \end{bmatrix} \nonumber
\end{align}
\subsection{Contact Model}
In order to avoid pre-specifying contact sequences, locations or timings, we need to enable the NMPC solver to reason about contacts. Some approaches resort to adding complementarity  constraints to enforce contacts (e.g.~\cite{posa2013complementary}). However, these constraints do not satisfy the Linear Independence Constraint Qualification (LICQ)~\cite{wrightLICQ}. Almost all off-the-shelf Nonlinear Optimal Control or Nonlinear Programming solvers assume LICQ and, therefore, cannot handle these problems~\cite{bouza2007mathematical}. In contrast, we add the contact physics to our dynamic model using an explicit contact model. As a result, the contact forces become an explicit function of the robot's state: $\vla(\vq,\dot{\vq}) = \vg(\vx(t))$.

Our contact model consists of a combination of linear springs and dampers perpendicular and parallel to the contact surface. For each end-effector we compute the contact model in the specialized contact frame~$C$ as follows:
\begin{align}
    {}_C\vla(\vq, \dot{\vq}) = 
        &-k \exp(\alpha_k~{}_C\vp_z(\vq)) \notag \\
        &-d~\sig(\alpha_d~{}_C\vp_z(\vq))~{}_C\dot{\vp}(\vq, \dot{\vq})
    \label{eq:contact_c}
\end{align}
with $d$ and $k$ being damper and spring parameters. In order to achieve smooth derivatives,
we multiply the damper-term with a sigmoid function of the normal component $\vp_z(\vq)$ of the contact surface penetration $\vp(\vq)$. Both the exponential and the sigmoid function serve as smoothing elements. Their `sharpness' is controlled by $\alpha_k$ and $\alpha_d$. Finally, the contact forces are transformed into the robot body frame by
\begin{align}
    {}_B\vla(\vq,\dot{\vq}) = \vR_{WB}(\vq) {}_C\vla(\vq,\dot{\vq})
    \label{eq:contact_b}
\end{align}
and subsequently passed to the forward dynamics where they act on the corresponding link. Our particular choice of the contact model smoothing supports the gradient-based solver by ensuring that contact forces never completely vanish. Therefore, the solver can reason about contacts even before they are established. While this is slightly nonphysical, we will later see that this does not hinder good performance on hardware. We emphasize that there exists physically more accurate explicit contact models \cite{azad2014new} and implicit contact models as popular in physics engines. The latter however rely on optimization based solvers which cannot be differentiated well. In contrast, our simplified model captures the governing effects accurately enough and allows for computing derivatives efficiently by using Auto-Diff which is key to solving the NMPC problem fast enough~\cite{Giftthaler2017autodiff}.

\section{NMPC APPROACH} 
\label{sub:solver}
Informally speaking, NMPC is achieved through solving the Nonlinear Optimal Control (NLOC) problem at sufficiently high rates. Popular approaches to nonlinear optimal control are Single Shooting, Multiple Shooting or Direct Collocation~\cite{diehl2006fast}, which discretize the continuous time optimal control problem and transcribe it into a Nonlinear Program (NLP). These NLPs are often solved using general off-the-shelf NLP solvers as presented in ~\cite{posa2016optimization, pardo2016evaluating}. However, such an approach does not fully exploit the sparsity structure inherent to optimal control problems and often results in poor algorithm runtimes, which are not fast enough for MPC applications.

To overcome this issue, we formulate our problem as an unconstrained optimal control problem, and resort to an optimized, custom solver, that implements a family of iterative Gauss-Newton NLOC algorithms~\cite{giftthaler2017family}. The solver employs a first-order method that locally approximates the NLOC problem as a Linear Quadratic Optimal Control (LQOC) problem using a Gauss-Newton Hessian approximation. The LQOC is solved by a Riccati-based solver which has linear complexity in the time horizon. This makes the approach efficient for larger time horizons. Our solver can be considered a generalization of the well-known iLQR~\cite{todorov2005ilqg} and SLQ~\cite{slq:2005} algorithms and covers both Single and Multiple Shooting. It designs time-varying state-feedback controllers of the form
\begin{equation}
\vu_n (\vx) = \vu^{ff}_n + \vK_n (\vx_n - \vx_n^\text{ref})
\label{eq:state_feedback_controller}
\end{equation}
where $\vu^{ff}_n$ is the feedforward control action and $\vK_n$ a linear feedback controller regulating deviations of the state $\vx_n$ from the reference trajectory $\vx_n^\text{ref}$. For most experiments in this paper, we use the iLQR algorithm. We furthermore compare it to the more efficient Gauss-Newton Multiple Shooting (GNMS) approach which can act as a direct replacement. Both algorithms use the same approach of formulating and solving a local LQOC problem, however their MPC formulations varies.

A summary of the iLQR-NMPC algorithm, is given in Algorithm~\ref{alg:mpc_algorithm}. It shows two forward integration steps during the algorithm. One directly after retrieving the state measurement to get the nominal trajectory, and a second one during the line search after updating the controller. For our application, the latter is important to obtain a new reference trajectory for the feedback controller to track. 
In contrast, the GNMS-NMPC algorithm, which is summarized in Algorithm~\ref{alg:gnms_mpc_algorithm}, designs a state reference trajectory simultaneously with the new control policy. 
Furthermore, it allows to separate the algorithm into a \emph{feedback} and a \emph{preparation} phase~\cite{diehl:2005:real}, which helps to minimize the latency between state-measurement and control policy update. For a detailed overview about the GNMS algorithm, the reader is referred to~\cite{giftthaler2017family}. An open-source reference implementation is available in~\cite{adrlCT}.
\begin{algorithm}[tpb] 
\caption{Discrete-time iLQR-MPC Algorithm} 
\label{alg:mpc_algorithm}
\begin{algorithmic} 
\scriptsize 
\STATE \textbf{Given}
\STATE - cost function~\eqref{eq:cost_function} and system dynamics~\eqref{eq:dynamic_system}.
\STATE - receding MPC time horizon $N$.
\STATE - stable initial control policy $\vu_n(\vx)$ of form~\eqref{eq:state_feedback_controller}
\STATE \textbf{Repeat Online:}
\STATE - get state measurement $\vx_\text{meas}$.
\STATE - forward integrate system dynamics~\eqref{eq:dynamic_system} with \mbox{$\vx_0 = \vx_\text{meas}$} to obtain 
\STATE \hspace{1em}state trajectories \mbox{$\vX = \{ \vx_0, \vx_1, \ldots, \vx_N\}$}, control trajectories
\STATE \hspace{1em}\mbox{$\vU = \{ \vu_0, \vu_1, \ldots, \vu_{N-1}\}$}
and corresponding sensitivities $\vA_n, \ \vB_n$.
\STATE - quadratize cost function~\eqref{eq:cost_function} around $\vX$ and $\vU$
\STATE - solve LQOC problem using a Riccati backward sweep
\STATE - retrieve control policy $\vu_n^{+}(\vx)$ of form~\eqref{eq:state_feedback_controller}
\STATE - \textbf{line search} over the control increment \mbox{$(\vu_n(\vx)^{+} - \vu_n(\vx))$}
\STATE \hspace{1em}and update $\vX^+$ by means of a forward simulation of the nonlinear
\STATE \hspace{1em}dynamics~\eqref{eq:dynamic_system} with \mbox{$\vx_0 = \vx_\text{meas}$}
\STATE - send policy $\vu_n^{+}(\vx)$ and $\vX^+$ to the robot tracking controller
\STATE - update $\vu_n(\vx) \leftarrow \vu_n^{+}(\vx)$
\end{algorithmic} 
\end{algorithm}

\begin{algorithm}[tpb] 
\caption{Discrete-time GNMS-NMPC Algorithm} 
\label{alg:gnms_mpc_algorithm}
\begin{algorithmic} 
\scriptsize 
\STATE \textbf{Given}
\STATE - cost function~\eqref{eq:cost_function} and system dynamics~\eqref{eq:dynamic_system}.
\STATE - receding MPC time horizon $N$.
\STATE - initial state and control trajectories \mbox{$\vX = \{ \vx_0, \vx_1, \ldots, \vx_N\}$},
\STATE \hspace{1em}\mbox{$\vU = \{ \vu_0, \vu_1, \ldots, \vu_{N-1}\}$}
\STATE \textbf{Repeat Online:}
\STATE \textit{Feedback phase}
\STATE - get state measurement $\vx_\text{meas}$.
\STATE - forward integrate system dynamics~\eqref{eq:dynamic_system} with \mbox{$\vx_0 = \vx_\text{meas}$} on the first 
\STATE \hspace{1em}multiple-shooting interval, obtain sensitivities $\vA_0, \ \vB_0$.
\STATE - quadratize cost function~\eqref{eq:cost_function} around $\vX$ and $\vU$ for first control stage
\STATE - solve LQOC problem using a Riccati backward sweep
\STATE - retrieve updated control policy $\vu_n^{+}(\vx)$ and updated trajectories $\vU^+$, $\vX^+$.
\STATE - send policy $\vu_n^{+}(\vx)$ and $\vX^+$ to the robot tracking controller
\STATE \textit{Preparation phase}
\STATE - update: $\vu_n(\vx) \leftarrow \vu_n^{+}(\vx)$, $\vX \leftarrow \vX^+$, $\vU \leftarrow \vU^+$
\STATE - forward integrate system dynamics~\eqref{eq:dynamic_system} for the multiple-shooting
\STATE \hspace{1em}intervals 1 to $N$, obtain sensitivities $\vA_1,\ldots \vA_{N-1}$, $\vB_1,\ldots, \vB_{N-1}$.
\STATE - quadratize cost function~\eqref{eq:cost_function} around $\vX$, $\vU$ for multiple-shooting intervals 1 to~$N$.
\end{algorithmic} 
\end{algorithm}

\section{SOFTWARE IMPLEMENTATION} \label{sec:software}
Running NMPC for a high dimensional system in real-time remains a challenge despite the powerful consumer PCs available today.
While the development of processors with faster clock speed has stalled in recent years, processing power instead foremost grows due to higher computation core counts as well as vectorization. However, both parallel execution and vectorization cannot be leveraged automatically by standard compilers. Also, many computational routines such as integrating a differential equation over time, are naturally sequential operations that cannot be parallelized easily.
In this subsection we describe how we optimize the NMPC solver from a numerical point of view and leverage the processor architecture to reduce the computational burden.

\subsection{Modelling Framework}
Our NMPC controller relies heavily on evaluating Rigid Body Dynamics and Kinematics. Therefore, we use RobCoGen~\cite{frigerioCodeGen}, an efficient code generation framework for modelling Rigid Body Dynamics. In~\cite{Giftthaler2017autodiff} we augmented RobCoGen to be compatible with Auto-Differentiation as implemented in our Control Toolbox (CT)~\cite{adrlCT}.  Furthermore, we use CT's contact model and kinematics that wrap around RobCoGen to provide contact force mappings. The resulting framework is lean compared to sophisticated physics engines and produces fast to evaluate, hard realtime capable code.

\subsection{Integration and Sensitivity Computation}
Our system dynamics include a contact model that needs to be chosen stiff enough to approximate the real physics of contact well. Naturally, this leads to a numerical stiffness which requires us to take small integration time-steps. Therefore, instead of using standard explicit integrators such as Euler or Runge-Kutta schemes, we use symplectic (or semi-implicit) integrators, which are also popular in physics engines~\cite{cattoSoftConstraints}. A symplectic integrator alternates between integrating positions and velocities, i.e. the updated positions are used to compute the velocity update and vice versa. Therefore, the computational complexity is similar to an explicit scheme, but the numerical stability is increased. In contrast to implicit integrators, we do not have to compute Jacobians explicitly and do not have to solve an internal optimization problem. The symplectic integrator allows us to increase the integration step size by a factor of four compared to explicit schemes~\cite{neunert:2017:ral}. 

Our LQOC solver requires us to compute sensitivities along the trajectory, i.e. partial derivatives of the integrated state with respect to the start state and control action of the step. These sensitivities can be understood as matrices $\vA_n$ and $\vB_n$ in a local linear approximation of the form $\vx_{n+1} = \vA_n \vx_n + \vB_n \vu_n + \vc_n$. While many existing approaches compute the sensitivities numerically~\cite{koenemann2015whole}, it is slow and can lead to severe numerical problems hindering convergence~\cite{Diehl2017}. Therefore, we compute them exactly by integrating a corresponding sensitivity ODE. A description of the integration scheme for sensitivities of symplectic integrators can be found in~\cite{prabhakar2015ergodic}. The linearization of our dynamic system is performed with Auto-Diff and code-generation described in detail in~\cite{Giftthaler2017autodiff}, which provides the same accuracy as analytic derivatives but outperforms them in terms of computational speed.

\subsection{Multithreading and Vectorization}
Another important factor for obtaining best performance is multi-threading. Some parts of our algorithm are inherently sequential, for example solving the LQOC problem backwards in time, or the forward simulation of the system when employing iLQR. However, using a multiple-shooting approach allows us to parallelize the forward simulation over the individual multiple-shooting intervals.
In this case, computation time decreases linearly with the number of cores. In contrast, iLQR requires to compute a single, continuous forward simulation and thus does not benefit from a multi-core processor in this step. The cost and sensitivity computation, which can be distributed among all available cores, is parallelizable for all our algorithm variants. 

Another optimization possibility is to use the processor's vectorization capabilities, which are Single Instruction Multiple Data (SIMD) implementations, especially useful for mathematical operations on vectors and matrices. In a previous implementation \cite{neunert:2017:ral}, we had already used SSE \cite{firasta2008intel} instructions. In this implementation, we switched to AVX~\cite{firasta2008intel} instructions. Since our code mostly consists of matrix and vector manipulations and register sizes of AVX are doubled over SSE, we obtained an additional speedup of almost a factor of two. This number is also supported by the release notes of the Eigen library~\cite{eigenweb} used for all computations. With the release of AVX512 doubling register sizes yet again, we expect another speedup of about factor of two.

\section{HARDWARE INTEGRATION}
\label{sec:hardware_integration}
\subsection{Platform Descriptions}
For hardware experiments we use two different quadruped robots, that strongly vary in size, weight and actuation principles. Therefore, the experiments underline the generality of the approach and show that no platform specific modifications are required. HyQ is an 80~kg, hydraulically actuated quadruped, while ANYmal weighs around 34~kg and uses electric, series-elastic actuators. Both robots are briefly described in the following.

\subsubsection{HyQ}
\label{sec:hyq_description}
HyQ~\cite{semini:2011:hyqjournal} is a fully torque controlled, hydraulic quadruped built by the Italian Institute of Technology. It features three joints per leg, namely Hip Abduction Adduction (HAA), Hip Flexion Extension (HFE) and Knee Flexion Extension (KFE). All joints are equipped with absolute and relative encoders, the joint torques are measured by load cells.

\subsubsection{ANYmal}
\label{sec:anymal_description}
While being more compact, ANYmal~\cite{hutter2016anymal} features the same joint configuration as HyQ. Furthermore, it is fully torque controlled via series-elastic actuators. Joint encoders before and after the elastic element measure its deflection which is used to compute the internal joint torque.

\subsection{Tracking Controller} 
\label{sub:tracking}
\begin{figure}[tbp]
\centering
\includegraphics[width=0.95\columnwidth]{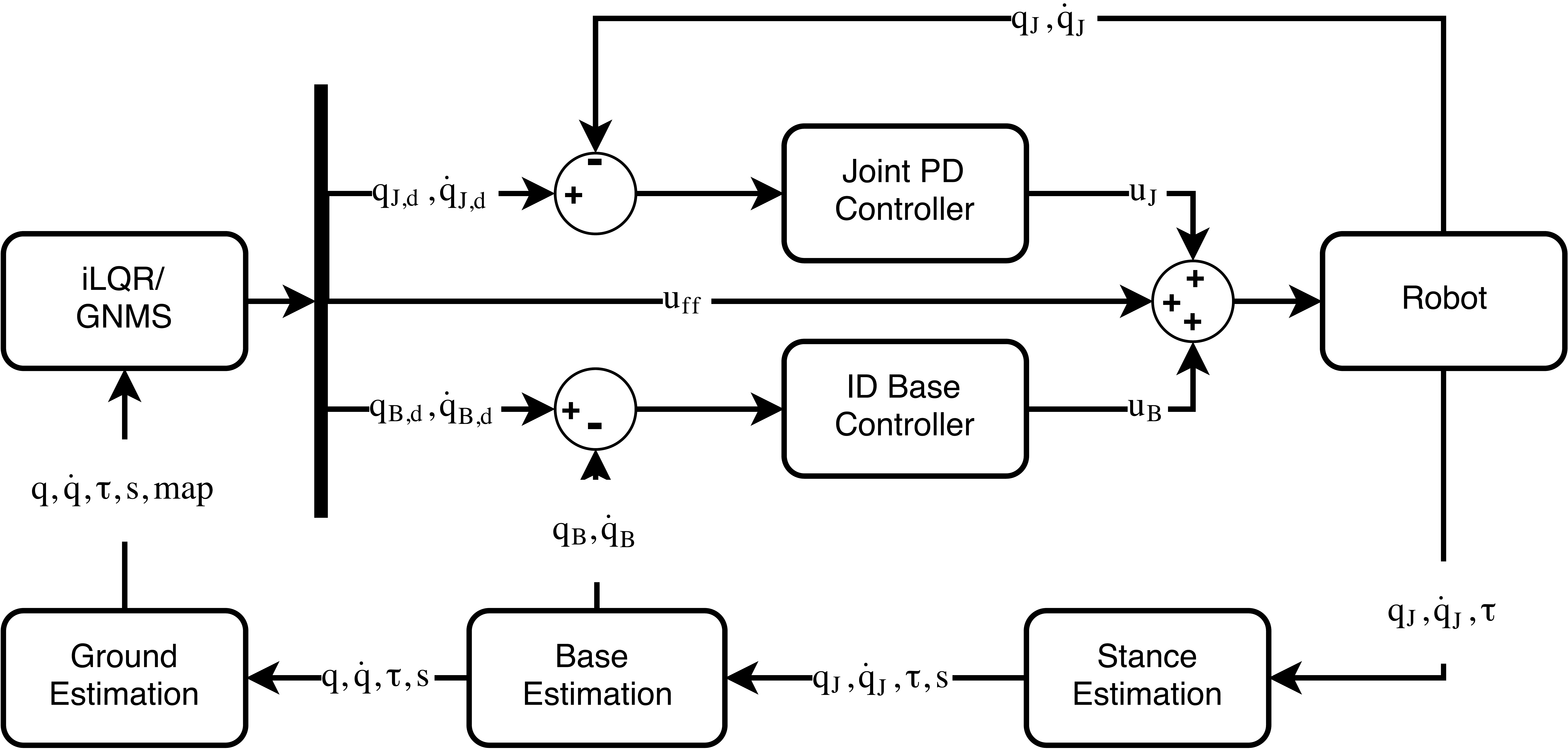}
\footnotesize
\caption{Structure of the estimation and control approach for hardware execution of the NMPC controller. Estimators estimate ground height and base state information. The optimized control input obtained from the NMPC solver is then augmented with the output of two tracking controllers.}
\label{fig:ctrl_structure}
\end{figure}

The control and tracking framework mostly corresponds to the one shown in Figure~\ref{fig:ctrl_structure} also used in \cite{neunert:2017:ral}. When running our NMPC framework, we directly apply the optimized torques as feedforward signal to the actuators. However, in our framework, there is an average delay of about 10-15~ms between state measurement and execution of the corresponding, optimized policy on the robot. This delay prevents from robustly controlling positions and velocities of the swing legs. Therefore, we add a PD control loop around the optimized joint trajectories that improves position and velocity tracking. Lastly, we also add a virtual model controller~\cite{pratt2001virtual} using inverse dynamics (without gravity compensation) which only contributes a few percent of the overall control signal. Therefore, the presented tasks also work without the base controller but show slightly improved performance with the controller enabled.

\subsection{State Estimation}
Both quadrupeds run a state estimator that fuses measurements from an Inertial Measurement Unit (IMU) and the leg encoders to estimate the base pose and twist of the robot. The state estimator is described in~\cite{bloesch2012state}. Joint position measurements are directly obtained from both robots and are then numerically differentiated to obtain joint velocities. Since we use an explicit contact model, an estimation of the ground is required. Here we check the stance of each foot and fit a plane through all stance legs. This ground estimator could possibly be replaced in the future by a mapping approach that delivers differentiable height information.

\subsection{Computing Setup}
In our setup, we use a dedicated computer that runs the NMPC control loop. The NMPC-node receives the current state of the robot via the Robot Operating System (ROS) from the midlevel control computer that executes the tracking controller described in subsection~\ref{sub:tracking}. Due to different software and hardware architectures, this tracking controller runs at 250~Hz on HyQ and at 400~Hz on ANYmal. The midlevel controller then sends desired torque, position and velocity setpoints to a lowlevel torque controller. In return, it receives current state measurements and computes the base and ground state estimate. The lowlevel controller runs a torque tracking and a position PD controller at 1~kHz on HyQ and 2.5~kHz on ANYmal. This controller is implemented on embedded hard real-time systems on both robots.

\section{RESULTS} \label{sec:results}
\begin{figure}[tbp]
\centering
\includegraphics[width=\columnwidth]{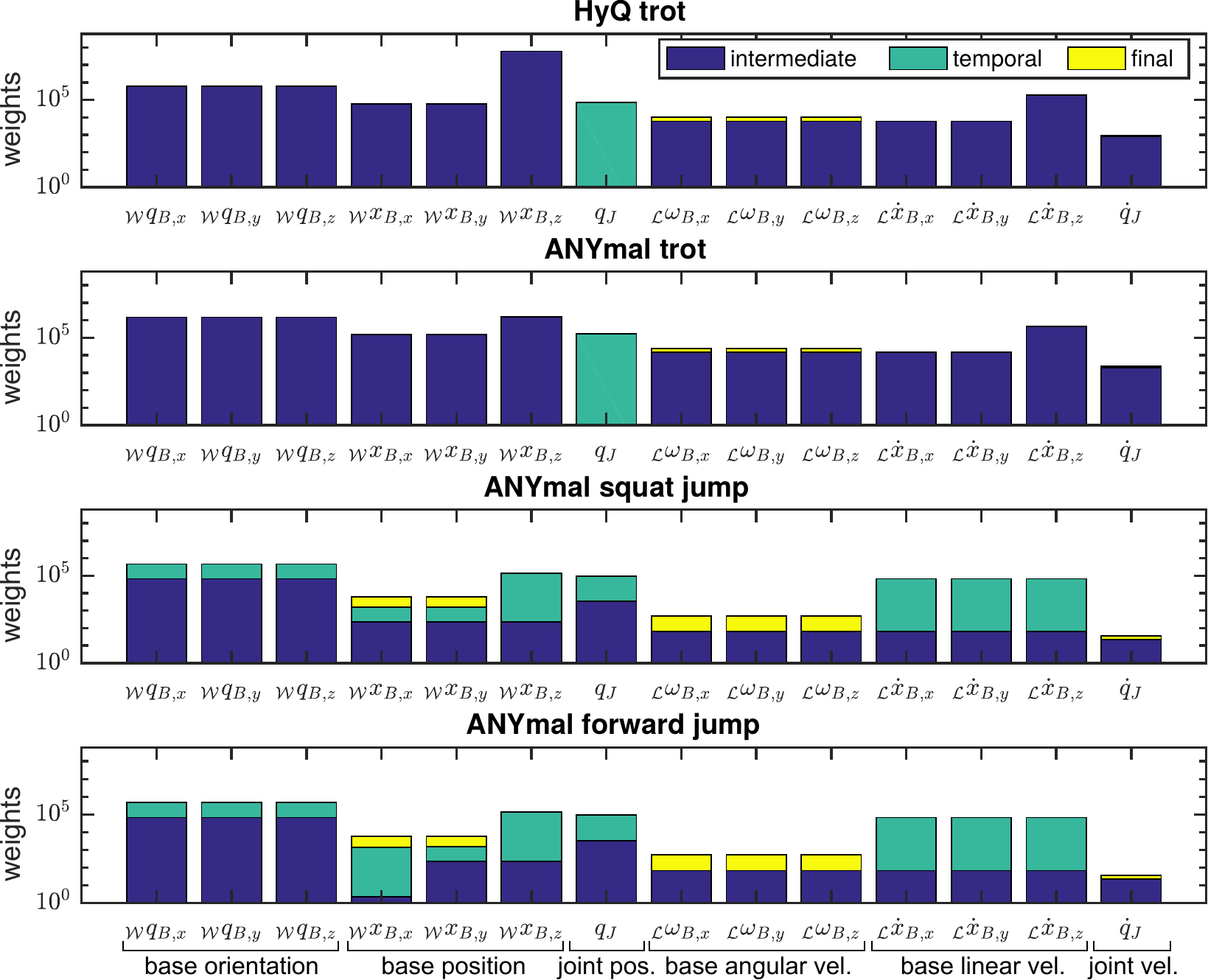}
\footnotesize
\caption{Cost function weights for the different hardware experiments plotted on a logarithmic scale. All weighting matrices are diagonal and thus the weights illustrated correspond to the diagonal entries. Weights on joint positions $q_J$ and velocities $\dot{q}_J$ are the same for all legs. The temporal costs in the trotting experiment affect swing leg pairs and are activated periodically. For the squat and forward jump temporal costs affect all legs equally and are activated once per jump.}
\label{fig:costFunctionWeights}
\end{figure}

The performance of our algorithms is assessed on both quadrupeds. We test a periodic trotting gait on both robots and disturb them during the tests. Furthermore, we execute additional dynamic motions on ANYmal. For all experiments, the cost functions weights are visualized in figure \ref{fig:costFunctionWeights} and provided as supplementary material.

In all experiments, we employ a time horizon of 500~ms, a control discretization of 4~ms and an integration rate of 1~ms for the aforementioned symplectic Euler scheme. In this work we employ the iLQR and GNMS solvers from~\cite{giftthaler2017family}, and use the solution from the previous MPC iteration as a warm start. We run a so called ``real-time iteration scheme''~\cite{diehl2006fast}, where we apply the optimized trajectory after a single iteration. Note that running only a single solver iteration before updating the state measurement results in better overall performance than running multiple iterations and letting the solver converge. All results are obtained with the same settings, solver and model. The different behaviors are thus simply a result of the choice of cost function and weights, offering a great versatility.

\subsection{Hardware Experiments HyQ}
\subsubsection{Trotting}
\begin{figure}[tbp]
\centering
\includegraphics[width=0.95\columnwidth]{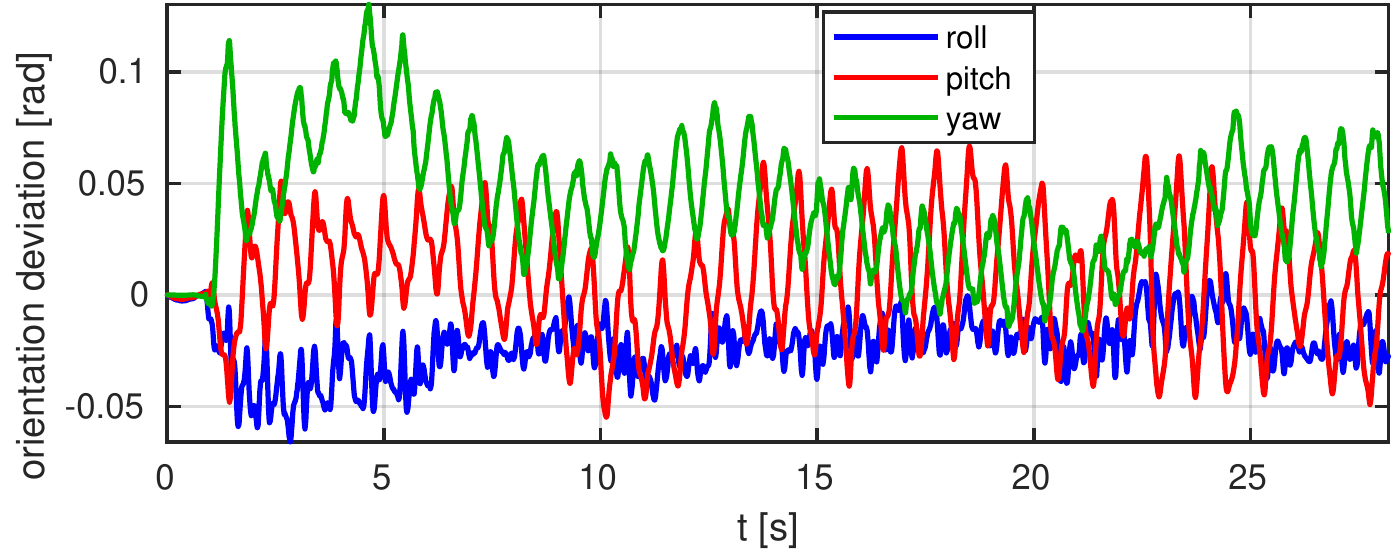}\\
\vspace{-0.3cm}
\includegraphics[width=0.95\columnwidth]{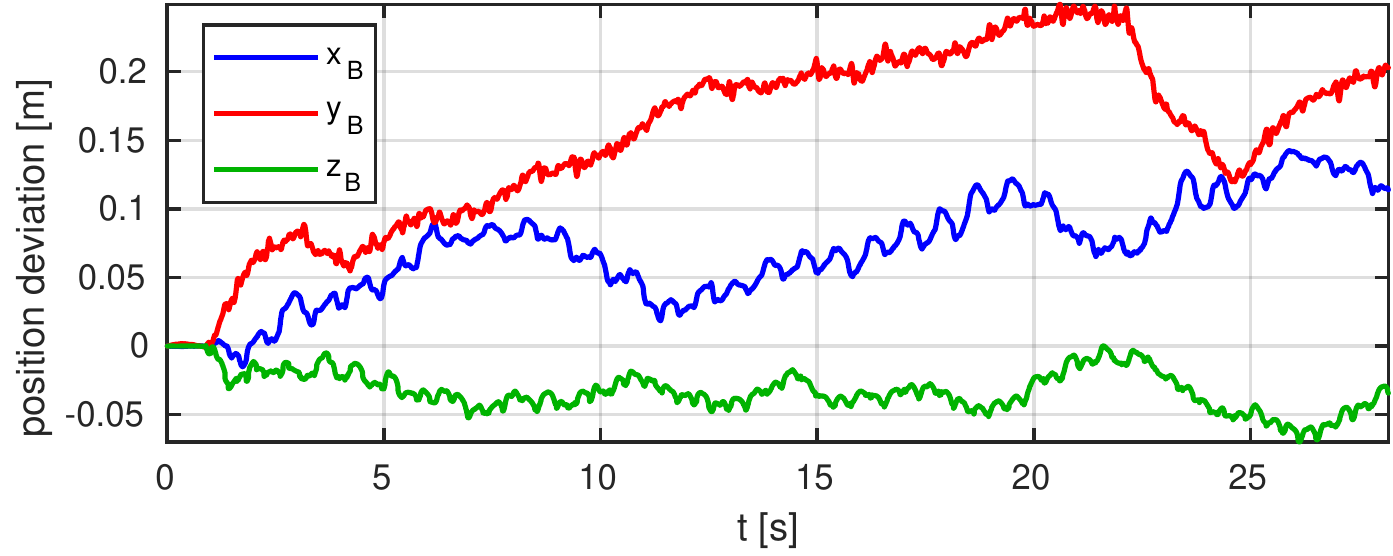}
\footnotesize
\caption{Plots of the deviation of the orientation and position offset during trotting experiments on HyQ. In the cost function, we penalize orientation stronger than position which shows in the plots. Compared to ANYmal the magnitude of the deviations is slightly larger. However, this is in part also due to the different sizes of both robots.}
\label{fig:trotDisturbanceHyQRot}
\end{figure}

As a first test, we investigate a periodic gait pattern, encouraged over periodically activated costs penalizing joint angle deviations from a desired swing leg apex configuration. In previous work~\cite{neunert:2017:ral}, we have demonstrated that our approach can also discover a trotting gait without swing leg costs. However, by adding such costs, we can influence the gait frequency and encourage trotting while staying in place. Since we only penalize the apex height of the swing leg, exact contact timings are not specified but can be adjusted by the algorithm. Additionally, since the swing phase is not a constraint, it can be violated by the algorithm if helpful for the overall performance. This can be observed during execution. If the robot base is strongly pushed to one side, the feet on that side are not lifted but remain on the ground to first stabilize the base. This means the algorithm naturally modifies the gait pattern and contact timings to optimize performance. Another observation is that, due to a feedforward dominant controller that plans ahead of time instead of simply reacting aggressively to disturbances, we obtain a compliant controller. HyQ can be perturbed significantly both on the base and the legs without reacting stiffly. Even placing planks under single feet does not deteriorate performance. Figure~\ref{fig:trotDisturbanceHyQRot} shows the results of a trotting experiment on HyQ. The plot shows that the base orientation and position deviations from the set point are regulated and remain small. Also, we add a strong cost penalty on the base orientation to improve stability. The resulting overall controller is stable and can robustly handle aforementioned disturbances.

\subsection{Hardware Experiments ANYmal}
\subsubsection{Trotting}
\begin{figure}[tbp]
\centering
\includegraphics[width=0.95\columnwidth]{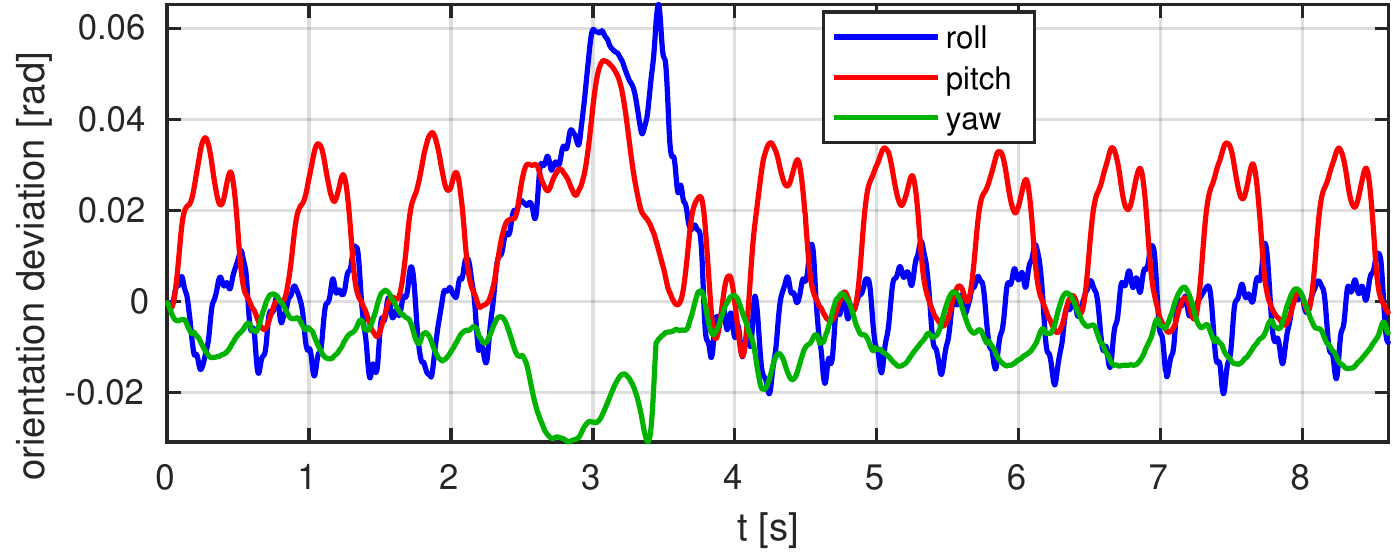}\\
\vspace{-0.3cm}
\includegraphics[width=0.95\columnwidth]{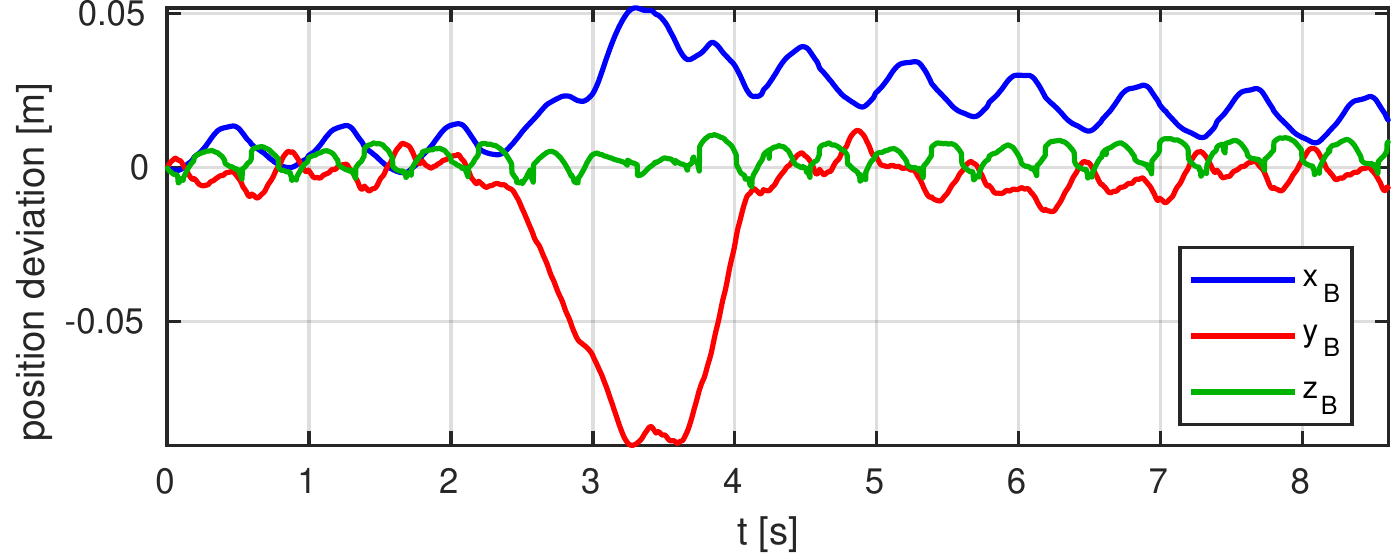}
\footnotesize
\caption{Base orientation (top) and position (bottom) of ANYmal during a trot. At t = 2.8s we placed a wooden stick below one foot which acted as heavy disturbance on the system. The controller is able to return to a periodic motion after the disturbance is removed. }
\label{fig:trotDisturbanceLinear}
\end{figure}

We repeat the same trotting experiments with adjusted weights (due to different mass/inertias) on ANYmal. Also here we observe that the controller is robust to disturbances. By adjusting the desired position and orientation of the base with a joystick, we navigate the robot around. Figure~\ref{fig:trotDisturbanceLinear} show measurements of the robot base during the trotting experiment. These plots illustrate how the MPC controller deals with disturbances. The robot is executing a periodic trot motion in place. At $t=2.8s$ we placed a wooden plank below one of ANYmal's feet. For the duration of the disturbance the MPC controller escapes the periodic trot and finds different solutions. To mitigate the disturbance the controller shifts and rotates the base and when the disturbance is gone it returns to the periodic trot. Compared to HyQ the deviations of the base orientation and position are much smaller. One of the reasons for this is certainly the size and weight differences between these two robots.

\subsubsection{Squat Jumps}
\begin{figure}[tbp]
\centering
\includegraphics[width=0.95\columnwidth]{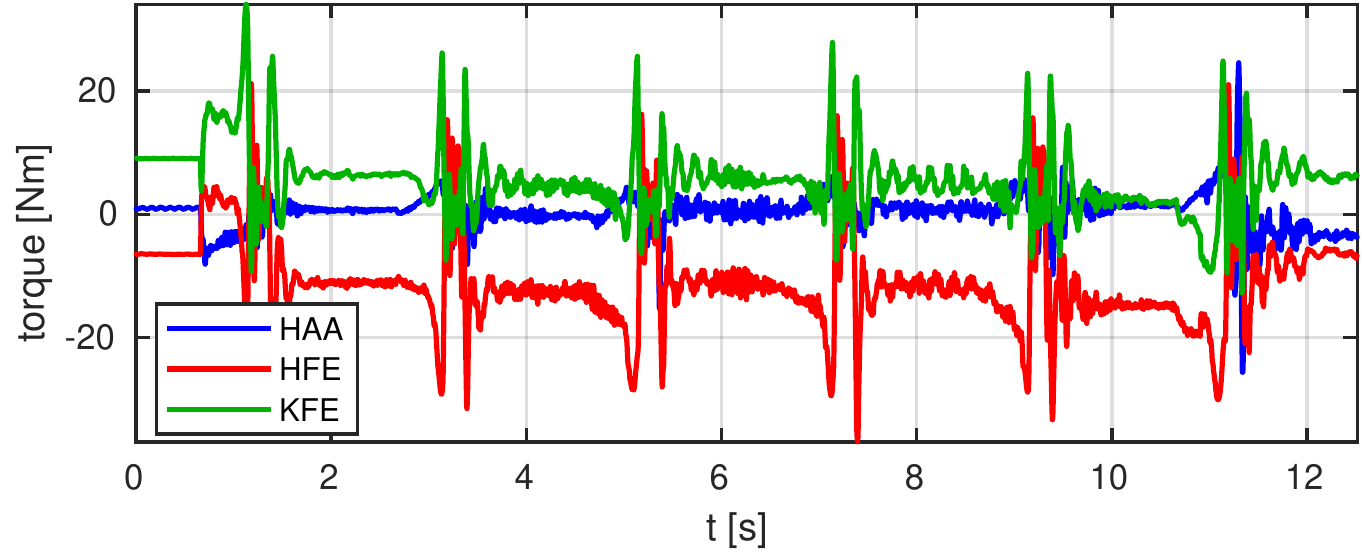}
\footnotesize
\caption{Joint torques of ANYmal during a repeated squat jump. The torques stay well below the physical torque limit of 40 Nm. Furthermore, the torques drop quite abruptly after take off and then peak during the landing phase.}
\label{fig:squatJumpJointTorques}
\end{figure}

\begin{figure}[tbp]
\centering
\includegraphics[width=0.95\columnwidth]{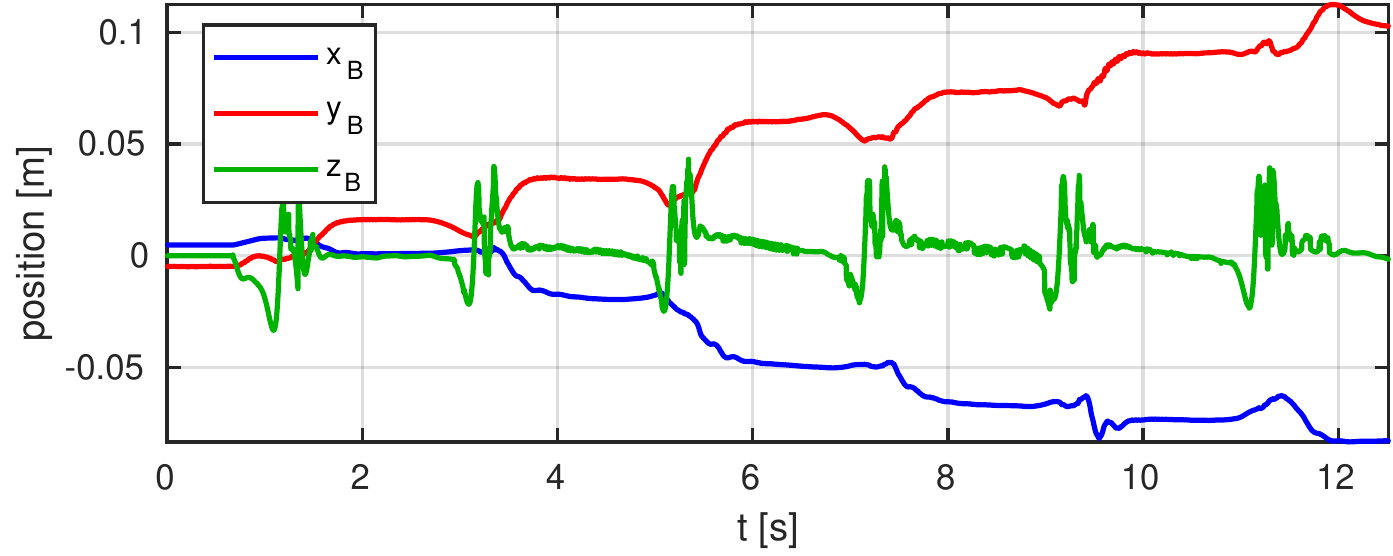}\\
\vspace{-0.3cm}
\includegraphics[width=0.935\columnwidth]{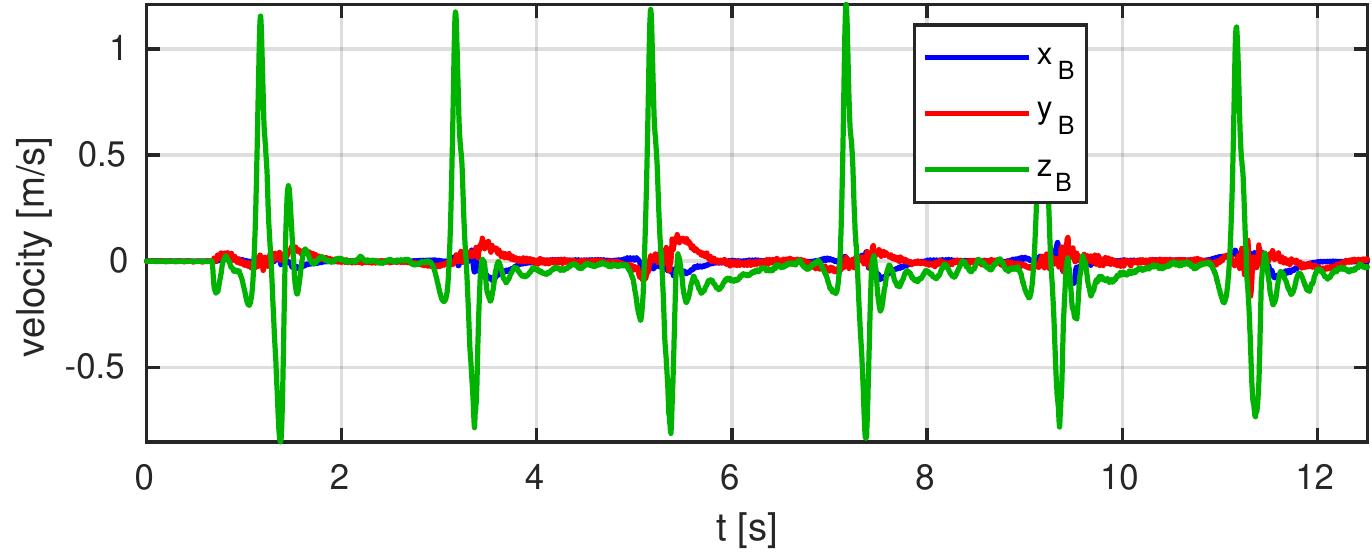}
\footnotesize
\caption{Base position and linear velocity of ANYmal during a repeated squat jump. We mostly penalize base orientation and linear velocities to obtain straight jump motions. As a result, the controller achieves a constant apex height but drifts slightly in x and y directions. We observe that ANYmal first slightly bents its knees and lowers the base before jumping off to accumulate energy, a behaviour that would not result from a pure feedback controller.}
\label{fig:squatJumpBasePos}
\end{figure}

To underline that the approach can leverage the full system dynamics, we also perform a squat jump. Here, amongst the usual running and final cost, a temporal cost encourages an upward base velocity at a certain time point. However, the lift-off time and especially the landing time are not pre-specified. Also, while it is not surprising that all legs lift-off at the same time, we do not explicitly enforce it. To test repeated jumps, we activate the vertical velocity costs periodically. While the robot does not always land perfectly, the MPC controller optimizes a trajectory from the current state and tries to get back as close as possible to the nominal state. This experiment underlines the robustness of the approach, as several squat jumps can be executed without resetting the robot to the nominal configuration.
Figures~\ref{fig:squatJumpBasePos} and \ref{fig:squatJumpJointTorques} show the measurements of ANYmal during the squat jump experiments. We enforce jumps every 2 seconds, starting at $t=1s$. The desired take off velocity was 1.0 m/s in vertical direction (z axis) leading to an apex height of 3 cm. The measurements show that the robot slightly overshoots the velocity setpoint by 0.1 m/s during execution. On the torque level the robot stayed well below the admissible torque level of ANYmal (40 Nm) without imposing additional constraints. The robot drifts in x and y direction between consecutive squat jumps. The reason for this is that we do not impose large cost penalties in x and y positions for stability reasons. One interesting observation is that the base height is lowered and the knees are bent before jumping off to accumulate kinetic energy, which reduces the maximum torque used for a given height.

\subsubsection{Forward Jump}
If we add a forward velocity component to the squat jump, we obtain a forward jump behavior. Again, lift-off time and landing time are not pre-specifed. Results of the forward jump can be found in the submitted video.

\subsection{Timings} 
\label{sub:timings}
In this section, we present timings for different solver setups. For all timings and experiments, the NMPC solver is run on an Intel Core i7 4790 quadcore PC with 3.6~GHz.
\begin{figure}[tbp]
\centering
\includegraphics[width=0.95\columnwidth]{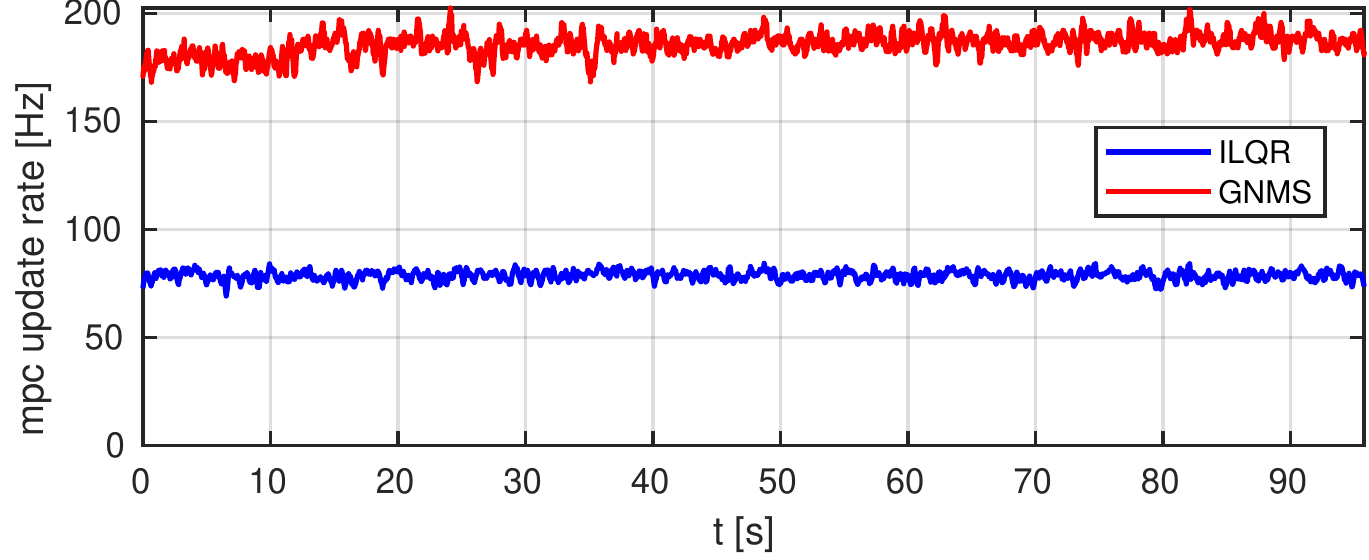}
\footnotesize
\caption{MPC update rate as recorded during two trotting experiments on ANYmal. While iLQR achieves update rates of around 80 Hz, GNMS reaches almost 190 Hz. While the higher update rate does not lead to significantly better performance, it leaves more headroom for extending the time horizon.}
\label{fig:timings}
\end{figure}

\begin{figure}[tbp]
\centering
\includegraphics[width=0.95\columnwidth]{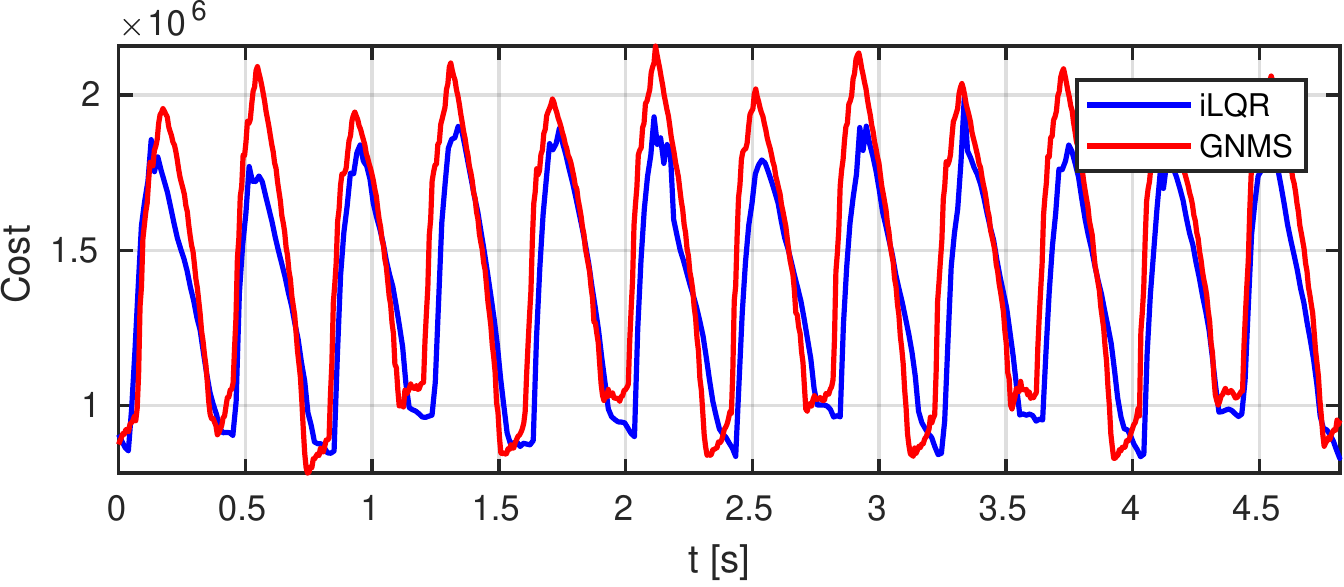}
\footnotesize
\caption{Total cost of the iLQR and GNMS algorithms during the trot experiment. Both algorithms result in a similar cost after optimization.}
\label{fig:costs}
\end{figure}

 Figure~\ref{fig:timings} shows the timings obtained from the trotting experiments on ANYmal. We measured the rate of incoming trajectories that the tracking controller received. It can be seen that our solver runs at around 80 Hz for when using the iLQR-algorithm and at 175 Hz when using Gauss-Newton Multiple-Shooting. The higher frequency of GNMS-MPC results from three main factors: first, the parallel forward simulation, second, in contrast to iLQR, GNMS does not require a line-search for this particular problem, and third, the GNMS algorithm allows for using a higher control discretization (6~ms). While GNMS offers a higher update rate, it is not very noticeable in performance such that iLQR performs similarly well. This is illustrated in Figure~\ref{fig:costs} which compares the costs of the trajectories obtained from both algorithms during a trot task. However, we notice that below 30 Hz update rate, the performance on hardware starts to degrade significantly. Therefore, GNMS leaves more headroom for more complex systems or longer time horizons.

At 80 Hz, a single computation takes less than 12~ms. Compared to~\cite{koenemann2015whole}, which is using the same algorithm but requires around 50 ms per iteration, our solver is about 300\% faster for the same time horizon and similar number of degrees of freedom while having a much finer control discretization (4~ms instead of 20~ms) and only using a third of the CPU cores. Assuming the same number of cores and the same discretization, our solver would run about 10-15 times faster than the one presented in~\cite{koenemann2015whole}. This results from a more efficient solver implementation, optimized vectorization, faster computation of the dynamics due to a simpler contact model as well as using Auto-Diff code-generation. As a result, delays are significantly reduced, which is crucial for robustness.

\section{SUMMARY AND OUTLOOK} \label{sec:summary}

The presented work shows the first application of whole-body NMPC through contacts for dynamic motions. Furthermore, this is the first time that such an approach is applied to hardware for periodic gait patterns and tasks with dynamic contact switches. We demonstrate that NMPC can be run at rates that exceed the state of the art by an order of magnitude by using an efficient implementation combined with state-of-the-art  software engineering techniques such as Auto-Diff, symplectic integration as well as vectorization. By applying our method to two different robots, we show that our integration with state estimation and controllers allows us to deploy the approach to different robots with different actuation principles without major adjustments. The tasks themselves underline the benefits of running an NMPC controller that can reason about contacts. When looking e.g. at the trotting task, we see disturbance-dependent gait pattern changes, base stabilization and small reactive side stepping. If implemented with a classical approach the resulting planning and control pipeline would possible consist of several modules and layers. Using NMPC they emerge naturally from a single algorithm. While cost function tuning remains a manual task to achieve optimal performance, new tasks can be simply tuned in a few minutes. This way, we avoid designing a control and planning framework from scratch when the task at hand changes, which is still often the case in robotics.

While these first results look promising, there is still a significant amount of future work to be considered, both from an algorithmic as well as from an implementation point of view. 
On the algorithmic side, we plan to leverage the speed advantage and better warm starting capabilities of GNMS to extend the time horizon of the NMPC controller. As part of this work, it would be worthwhile to see how a larger time horizon influences performance and robustness. We expect that a longer time horizon could show more elaborate disturbance rejection and recovery behavior since it offers more flexibility and predictive capabilities to the solver. Furthermore, while most tasks by design stayed within the physical limitations of the platforms, GNMS would allow us to handle constraints such as torque limitations explicitly as as inequality constraints. Given that the higher update rate of GNMS-MPC did not lead to an increase in performance, we believe that our method can cope with the extra computational complexity when dealing with constraints, especially since the algorithm's complexity remains linear in time. Also, we could resort to soft-constraints rather than hard constraints \cite{neunert_humanoids}.
From a practical perspective, we wish to include some model adaptation or disturbance estimation to our implementation that robustifies our approach even further. Additionally, this could help to bring new motions to hardware. Finally, the NMPC controller could also be used to track and stabilize motions generated from a high level foothold and motion planner to handle non-convex terrains.




\section*{\small{ACKNOWLEDGMENT}}
\footnotesize
We wish to thank the entire SYSCOP group at IMTEK at the University of Freiburg, especially Prof. Moritz Diehl and Dimitris Kouzoupis for their valuable feedback and Gianluca Frison for his support for interfacing with the HPIPM solver. Furthermore, our gratitude goes to Marko Bjelonic and P\'{e}ter Fankhauser who helped with the experiments.
This research is supported by the Swiss National Science Foundation through the NCCR Digital Fabrication, the NCCR Robotics and a Professorship Award to Jonas Buchli.

\footnotesize
\bibliographystyle{ieeetr}
\bibliography{refs}

\end{document}